\documentclass{article}

\usepackage{microtype}
\usepackage{graphicx}
\usepackage{booktabs} % for professional tables

\usepackage{hyperref}
\usepackage{amsmath}
\usepackage{amsfonts}
\usepackage{subfig}
\usepackage{enumitem}
\usepackage{float}

\newcommand{\e}{\epsilon}
\newcommand{\y}{\gamma}
\newcommand{\s}{\sigma}
\newcommand{\E}{\mathbb{E}}
\newcommand{\N}{\mathcal{N}}
\newcommand{\lb}{\left [}
\newcommand{\rb}{\right ]}
\newcommand{\lp}{\left (}
\newcommand{\rp}{\right )}
\newcommand{\B}{\mathcal{B}}

\DeclareMathOperator*{\argmin}{argmin}
\DeclareMathOperator*{\argmax}{argmax}
\DeclareMathOperator*{\clip}{clip}

\usepackage[accepted]{icml2018}

\icmltitlerunning{Addressing Function Approximation Error in Actor-Critic Methods}

\begin{document}

\twocolumn[
\icmltitle{Addressing Function Approximation Error in Actor-Critic Methods}



\icmlsetsymbol{equal}{*}

\begin{icmlauthorlist}
\icmlauthor{Scott Fujimoto}{mcgill}
\icmlauthor{Herke van Hoof}{ams}
\icmlauthor{David Meger}{mcgill}
\end{icmlauthorlist}

\icmlaffiliation{mcgill}{McGill University, Montreal, Canada}
\icmlaffiliation{ams}{University of Amsterdam, Amsterdam, Netherlands}
\icmlcorrespondingauthor{Scott Fujimoto}{scott.fujimoto@mail.mcgill.ca}
\icmlkeywords{reinforcement learning, deep learning}

\vskip 0.3in
]



\printAffiliationsAndNotice{}  % leave blank if no need to mention equal contribution

\begin{abstract}
In value-based reinforcement learning methods such as deep Q-learning, function approximation errors are known to lead  to overestimated value estimates and suboptimal policies. We show that this problem persists in an actor-critic setting and propose novel mechanisms to minimize its effects on both the actor and the critic. Our algorithm builds on Double Q-learning, by taking the minimum value between a pair of critics to limit overestimation. We draw the connection between target networks and overestimation bias, and suggest delaying policy updates to reduce per-update error and further improve performance. We evaluate our method on the suite of OpenAI gym tasks, outperforming the state of the art in every environment tested.
\end{abstract}

%%%%%%%%%%%%%%%%%%%%%%%%%%%%%%%%%%%%%%%%%
\section{Introduction}
%%%%%%%%%%%%%%%%%%%%%%%%%%%%%%%%%%%%%%%%%

In reinforcement learning problems with discrete action spaces, the issue of value overestimation as a result of function approximation errors is well-studied. However, similar issues with actor-critic methods in continuous control domains have been largely left untouched. In this paper, we show overestimation bias and the accumulation of error in temporal difference methods are present in an actor-critic setting. Our proposed method addresses these issues, and greatly outperforms the current state of the art.

Overestimation bias is a property of Q-learning in which the maximization of a noisy value estimate induces a consistent overestimation \cite{thrun1993bias}. In a function approximation setting, this noise is unavoidable given the imprecision of the estimator. This inaccuracy is further exaggerated by the nature of temporal difference learning \cite{sutton1988tdlearning}, in which an estimate of the value function is updated using the estimate of a subsequent state. This means using an imprecise estimate within each update will lead to an accumulation of error.
Due to overestimation bias, this accumulated error can cause arbitrarily bad states to be estimated as high value, resulting in suboptimal policy updates and divergent behavior.

This paper begins by establishing this overestimation property is also present for deterministic policy gradients \cite{DPG}, in the continuous control setting. Furthermore, we find the ubiquitous solution in the discrete action setting, Double DQN \cite{DoubleDQN}, to be ineffective in an actor-critic setting. During training, Double DQN estimates the value of the current policy with a separate target value function, allowing actions to be evaluated without maximization bias. Unfortunately, due to the slow-changing policy in an actor-critic setting, the current and target value estimates remain too similar to avoid maximization bias. This can be dealt with by adapting an older variant, Double Q-learning \cite{hasselt2010double}, to an actor-critic format by using a pair of independently trained critics. While this allows for a less biased value estimation, even an unbiased estimate with high variance can still lead to future overestimations in local regions of state space, which in turn can negatively affect the global policy. To address this concern, we propose a clipped Double Q-learning variant which leverages the notion that a value estimate suffering from overestimation bias can be used as an approximate upper-bound to the true value estimate. This favors underestimations, which do not tend to be propagated during learning, as actions with low value estimates are avoided by the policy.

Given the connection of noise to overestimation bias, this paper contains a number of components that address variance reduction. First, we show that target networks, a common approach in deep Q-learning methods, are critical for variance reduction by reducing the accumulation of errors. Second, to address the coupling of value and policy, we propose delaying policy updates until the value estimate has converged. Finally, we introduce a novel regularization strategy, where a SARSA-style update bootstraps similar action estimates to further reduce variance. 

Our modifications are applied to the state of the art actor-critic method for continuous control, Deep Deterministic Policy Gradient algorithm (DDPG) \cite{DDPG}, to form the Twin Delayed Deep Deterministic policy gradient algorithm (TD3), an actor-critic algorithm which considers the interplay between function approximation error in both policy and value updates. We evaluate our algorithm on seven continuous control domains from OpenAI gym \cite{OpenAIGym}, where we outperform the state of the art by a wide margin.

Given the recent concerns in reproducibility \cite{hendersonRL2017}, we run our experiments across a large number of seeds with fair evaluation metrics, perform ablation studies across each contribution, and open source both our code and learning curves (\url{https://github.com/sfujim/TD3}). 

%%%%%%%%%%%%%%%%%%%%%%%%%%%%%%%%%%%%%%%%%
\section{Related Work}
%%%%%%%%%%%%%%%%%%%%%%%%%%%%%%%%%%%%%%%%%

Function approximation error and its effect on bias and variance in reinforcement learning algorithms have been studied in prior works \cite{pendrith1997estimator, mannor2007bias}. Our work focuses on two outcomes that occur as the result of estimation error, namely overestimation bias and a high variance build-up.  

Several approaches exist to reduce the effects of overestimation bias due to function approximation and policy optimization in Q-learning. Double Q-learning uses two independent estimators to make unbiased value estimates \cite{hasselt2010double, DoubleDQN}. Other approaches have focused directly on reducing the variance \cite{AVGDQN}, minimizing over-fitting to early high variance estimates \cite{fox2015glearning}, or through corrective terms \cite{lee2013biascor}. Further, the variance of the value estimate has been considered directly for risk-aversion \cite{mannor2011mean} and exploration \cite{UncertaintyBellman}, but without connection to overestimation bias. 

The concern of variance due to the accumulation of error in temporal difference learning has been largely dealt with by either minimizing the size of errors at each time step or mixing off-policy and Monte-Carlo returns. Our work shows the importance of a standard technique, target networks, for the reduction of per-update error, and develops a regularization technique for the variance reduction by averaging over value estimates. Concurrently, \citet{Smoothie} showed smoothed value functions could be used to train stochastic policies with reduced variance and improved performance. Methods with multi-step returns offer a trade-off between accumulated estimation bias and variance induced by the policy and the environment. These methods have been shown to be an effective approach, through importance sampling \cite{precup2001off, munos2016safe}, distributed methods \cite{a3c,espeholt2018impala}, and approximate bounds \cite{optimalitytightening}. However, rather than provide a direct solution to the accumulation of error, these methods circumvent the problem by considering a longer horizon. Another approach is a reduction in the discount factor \cite{petrik2009discount}, reducing the contribution of each error.

Our method builds on the Deterministic Policy Gradient algorithm (DPG) \citep{DPG}, an actor-critic method which uses a learned value estimate to train a deterministic policy. An extension of DPG to deep reinforcement learning, DDPG \cite{DDPG}, has shown to produce state of the art results with an efficient number of iterations.
Orthogonal to our approach, recent improvements to DDPG include distributed methods \cite{popov2017data}, along with multi-step returns and prioritized experience replay \cite{PrioritizedExpReplay, horgan2018distributed}, and distributional methods \cite{bellemare2017distributional, barth-maron2018distributional}. 
% In our experiments, we focus on the single-step return case, but the improvements we propose are compatible with the use of multi-step returns. 
%In contrast with our method, work has been done on adapting deep Q-learning methods for continuous actions, which replace selection of the action with the highest estimated value with various approximations \cite{gu2016continuous,metz2017discrete,tavakoli2017action}.

%%%%%%%%%%%%%%%%%%%%%%%%%%%%%%%%%%%%%%%%%
\section{Background}
%%%%%%%%%%%%%%%%%%%%%%%%%%%%%%%%%%%%%%%%%

Reinforcement learning considers the paradigm of an agent interacting with its environment with the aim of learning reward-maximizing behavior. At each discrete time step $t$, with a given state $s \in \mathcal{S}$, the agent selects actions $a \in \mathcal{A}$ with respect to its policy $\pi: \mathcal{S} \rightarrow \mathcal{A}$, receiving a reward $r$ and the new state of the environment $s'$. The return is defined as the discounted sum of rewards $R_t = \sum_{i=t}^{T} \y^{i-t} r(s_i, a_i)$, where $\y$ is a discount factor determining the priority of short-term rewards. 

In reinforcement learning, the objective is to find the optimal policy $\pi_\phi$, with parameters $\phi$, which maximizes the expected return $J(\phi) = \E_{s_i \sim p_\pi, a_i \sim \pi} \lb R_0 \rb$. For continuous control, parametrized policies $\pi_\phi$ can be updated by taking the gradient of the expected return $\nabla_{\phi} J(\phi)$. In actor-critic methods, the policy, known as the actor, can be updated through the deterministic policy gradient algorithm \cite{DPG}:
\begin{equation}
\nabla_{\phi} J(\phi) = \E_{s \sim p_\pi} \lb \nabla_a Q^\pi(s,a)|_{a=\pi(s)} \nabla_{\phi} \pi_\phi(s) \rb. 
\end{equation}
$Q^\pi(s,a) = \E_{s_i \sim p_\pi, a_i \sim \pi} \lb R_t | s, a \rb$, the expected return when performing action $a$ in state $s$ and following $\pi$ after, is known as the critic or the value function. 

In Q-learning, the value function can be learned using temporal difference learning \cite{sutton1988tdlearning,watkins1989qlearning}, an update rule based on the Bellman equation \cite{bellman}. The Bellman equation is a fundamental relationship between the value of a state-action pair $(s,a)$ and the value of the subsequent state-action pair $(s',a')$:
\begin{equation}
Q^\pi(s, a) = r + \y \E_{s', a'} \lb Q^\pi(s', a') \rb, \quad a' \sim \pi(s').
\end{equation}

For a large state space, the value can be estimated with a differentiable function approximator $Q_\theta(s, a)$, with parameters $\theta$. In deep Q-learning \cite{DQN}, the network is updated by using temporal difference learning with a secondary frozen target network $Q_{\theta'}(s,a)$ to maintain a fixed objective $y$ over multiple updates:
\begin{equation}
y = r + \y Q_{\theta'}(s',a'), \quad a' \sim \pi_{\phi'}(s'),
\end{equation}
where the actions are selected from a target actor network $\pi_{\phi'}$. The weights of a target network are either updated periodically to exactly match the weights of the current network, or by some proportion $\tau$ at each time step $\theta' \leftarrow \tau \theta + (1 - \tau) \theta'$. This update can be applied in an off-policy fashion, sampling random mini-batches of transitions from an experience replay buffer~\cite{expreplay1992}.

%%%%%%%%%%%%%%%%%%%%%%%%%%%%%%%%%%%%%%%%%
\section{Overestimation Bias} \label{sec:over}
%%%%%%%%%%%%%%%%%%%%%%%%%%%%%%%%%%%%%%%%%

In Q-learning with discrete actions, the value estimate is updated with a greedy target $y = r + \y \max_{a'} Q(s',a')$, however, if the target is susceptible to error $\e$, then the maximum over the value along with its error will generally be greater than the true maximum, $\E_\e[\max_{a'} (Q(s',a') + \e)] \geq \max_{a'} Q(s',a')$ \cite{thrun1993bias}. As a result, even initially zero-mean error can cause value updates to result in a consistent overestimation bias, which is then propagated through the Bellman equation. This is problematic as errors induced by function approximation are unavoidable. 

While in the discrete action setting overestimation bias is an obvious artifact from the analytical maximization, the presence and effects of overestimation bias is less clear in an actor-critic setting where the policy is updated via gradient descent. We begin by proving that the value estimate in deterministic policy gradients will be an overestimation under some basic assumptions in Section \ref{sec:overac} and then propose a clipped variant of Double Q-learning in an actor-critic setting to reduce overestimation bias in Section \ref{sec:cdq}. 

%%%%%%%%%%%%%%%%%%%%%%%%%%%%%%%%%%%%%%%%%
\subsection{Overestimation Bias in Actor-Critic} \label{sec:overac}
%%%%%%%%%%%%%%%%%%%%%%%%%%%%%%%%%%%%%%%%%

In actor-critic methods the policy is updated with respect to the value estimates of an approximate critic. In this section we assume the policy is updated using the deterministic policy gradient, and show that the update induces overestimation in the value estimate. 
Given current policy parameters $\phi$, let $\phi_\textnormal{approx}$ define the parameters from the actor update induced by the maximization of the approximate critic $Q_\theta(s,a)$ and $\phi_\textnormal{true}$ the parameters from the hypothetical actor update with respect to the true underlying value function $Q^\pi(s,a)$ (which is not known during learning): 
\begin{equation}
\begin{aligned}
\phi_\textnormal{approx} &= \phi + \frac{\alpha}{Z_1} \E_{s \sim p_\pi} \lb \nabla_\phi \pi_\phi(s) \nabla_a Q_\theta(s,a)|_{a=\pi_\phi(s)} \rb\\
\phi_\textnormal{true} &= \phi + \frac{\alpha}{Z_2} \E_{s \sim p_\pi} \lb \nabla_\phi \pi_\phi(s) \nabla_a Q^\pi(s,a)|_{a=\pi_\phi(s)} \rb,
\end{aligned}
\end{equation} 
where we assume $Z_1$ and $Z_2$ are chosen to normalize the gradient, i.e., such that $Z^{-1}||\E[\cdot]|| = 1$. Without normalized gradients, overestimation bias is still guaranteed to occur with slightly stricter conditions. We examine this case further in the supplementary material. We denote $\pi_\textnormal{approx}$ and $\pi_\textnormal{true}$ as the policy with parameters $\phi_\textnormal{approx}$ and $\phi_\textnormal{true}$ respectively. 

As the gradient direction is a local maximizer, there exists $\e_1$ sufficiently small such that if $\alpha \leq \e_1$ then the \emph{approximate} value of $\pi_\textnormal{approx}$ will be bounded below by the \emph{approximate} value of $\pi_\textnormal{true}$:
\begin{equation} \label{approxQ}
\E \lb Q_\theta(s, \pi_\textnormal{approx}(s)) \rb 
\geq 
\E \lb Q_\theta(s, \pi_\textnormal{true}(s)) \rb.
\end{equation}
Conversely, there exists $\e_2$ sufficiently small such that if $\alpha \leq \e_2$ then the \emph{true} value of $\pi_\textnormal{approx}$ will be bounded above by the \emph{true} value of $\pi_\textnormal{true}$:
\begin{equation} \label{trueQ}
\E \lb Q^\pi(s, \pi_\textnormal{true}(s)) \rb
\geq 
\E \lb Q^\pi(s, \pi_\textnormal{approx}(s)) \rb.
\end{equation}
If in expectation the value estimate is at least as large as the \textit{true} value with respect to $\phi_{true}$, 
$\E \lb Q_\theta \lp s, \pi_\textnormal{true}(s) \rp \rb \geq \E \lb Q^\pi \lp s, \pi_\textnormal{true}(s) \rp \rb$, then Equations (\ref{approxQ}) and (\ref{trueQ}) imply that if $\alpha < \min(\e_1, \e_2)$, then the value estimate will be overestimated:
\begin{equation}
\label{overestimation}
\E \lb Q_\theta(s, \pi_\textnormal{approx}(s)) \rb
\geq 
\E \lb Q^\pi(s, \pi_\textnormal{approx}(s)) \rb. 
\end{equation}

Although this overestimation may be minimal with each update, the presence of error raises two concerns. Firstly, the overestimation may develop into a more significant bias over many updates if left unchecked. Secondly, an inaccurate value estimate may lead to poor policy updates. This is particularly problematic because a feedback loop is created, in which suboptimal actions might be highly rated by the suboptimal critic, reinforcing the suboptimal action in the next policy update.

\begin{figure} 
\centering
\captionsetup[subfloat]{captionskip=-8pt}
\includegraphics[width=\linewidth]{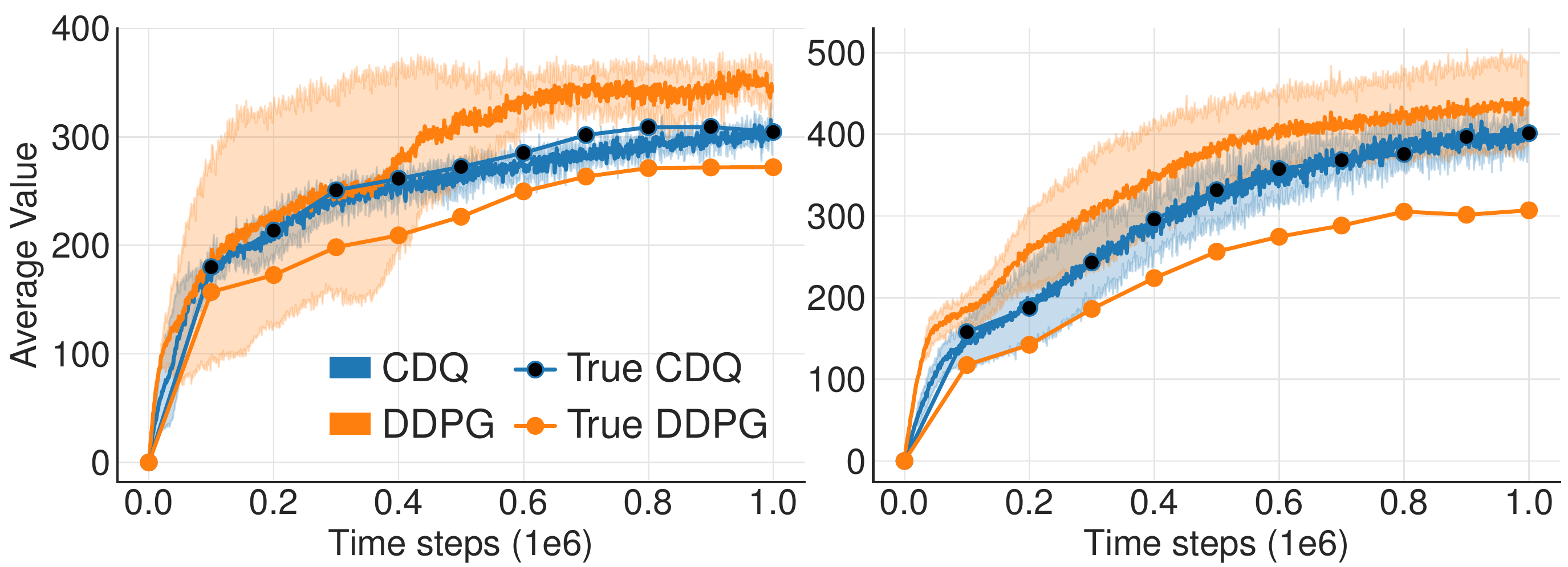}
\subfloat[Hopper-v1]{\hspace{0.56\linewidth}}
\subfloat[Walker2d-v1]{\hspace{0.44\linewidth}}
\caption{Measuring overestimation bias in the value estimates of DDPG and our proposed method, Clipped Double Q-learning (CDQ), on MuJoCo environments over 1 million time steps.} % 3 trials 
\label{fig:overestimation}
\end{figure}

\textbf{Does this theoretical overestimation occur in practice for state-of-the-art methods?} We answer this question by plotting the value estimate of DDPG \cite{DDPG} over time while it learns on the OpenAI gym environments Hopper-v1 and Walker2d-v1 \cite{OpenAIGym}. In Figure~\ref{fig:overestimation}, we graph the average value estimate over 10000 states and compare it to an estimate of the true value. The true value is estimated using the average discounted return over 1000 episodes following the current policy, starting from states sampled from the replay buffer. A very clear overestimation bias occurs from the learning procedure, which contrasts with the novel method that we describe in the following section, Clipped Double Q-learning, which greatly reduces overestimation by the critic. 

%%%%%%%%%%%%%%%%%%%%%%%%%%%%%%%%%%%%%%%%%
\subsection{Clipped Double Q-Learning for Actor-Critic} \label{sec:cdq}
%%%%%%%%%%%%%%%%%%%%%%%%%%%%%%%%%%%%%%%%%

\begin{figure} 
\centering
\captionsetup[subfloat]{captionskip=-8pt}
\includegraphics[width=\linewidth]{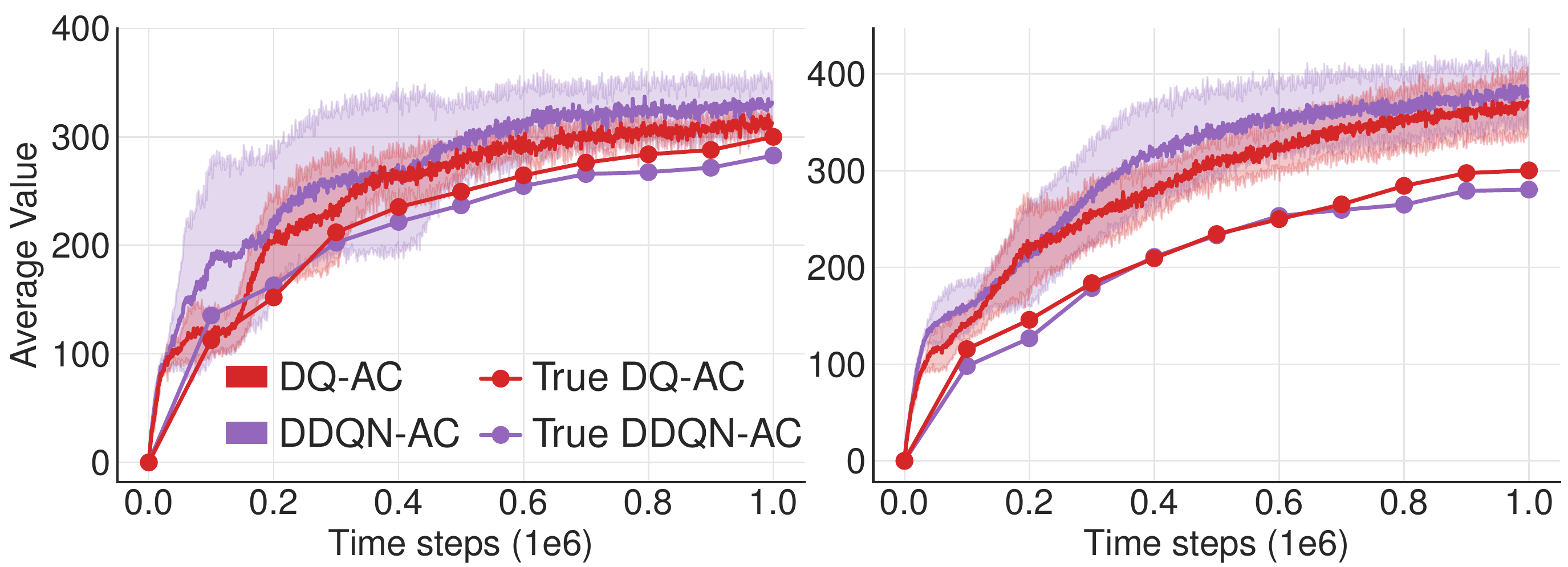}
\subfloat[Hopper-v1]{\hspace{0.56\linewidth}}
\subfloat[Walker2d-v1]{\hspace{0.44\linewidth}}
\caption{Measuring overestimation bias in the value estimates of actor critic variants of Double DQN (DDQN-AC) and Double Q-learning (DQ-AC) on MuJoCo environments over 1 million time steps.}
% 3 trials 
\label{fig:doubleestimation}
\end{figure}

While several approaches to reducing overestimation bias have been proposed, we find them ineffective in an actor-critic setting. This section introduces a novel clipped variant of Double Q-learning \cite{hasselt2010double}, which can replace the critic in any actor-critic method. 

In Double Q-learning, the greedy update is disentangled from the value function by maintaining two separate value estimates, each of which is used to update the other. If the value estimates are independent, they can be used to make unbiased estimates of the actions selected using the opposite value estimate. In Double DQN \cite{DoubleDQN}, the authors propose using the target network as one of the value estimates, and obtain a policy by greedy maximization of the current value network rather than the target network. In an actor-critic setting, an analogous update uses the current policy rather than the target policy in the learning target:
\begin{equation}
y = r + \y Q_{\theta'}(s', \pi_\phi(s')).
\end{equation}
% and fast updating target network ? 
In practice however, we found that with the slow-changing policy in actor-critic, the current and target networks were too similar to make an independent estimation, and offered little improvement.
%over the original learning target. 
Instead, the original Double Q-learning formulation can be used, with a pair of actors ($\pi_{\phi_1}$, $\pi_{\phi_2}$) and critics ($Q_{\theta_1}$, $Q_{\theta_2}$), where $\pi_{\phi_1}$ is optimized with respect to $Q_{\theta_1}$ and $\pi_{\phi_2}$ with respect to $Q_{\theta_2}$:
\begin{equation}
\begin{aligned}
y_1 &= r + \y Q_{\theta'_2}(s',\pi_{\phi_1}(s')) \\
y_2 &= r + \y Q_{\theta'_1}(s',\pi_{\phi_2}(s')). \\
\end{aligned}
\end{equation}
We measure the overestimation bias in Figure \ref{fig:doubleestimation}, which demonstrates that the actor-critic Double DQN suffers from a similar overestimation as DDPG (as shown in Figure \ref{fig:overestimation}). While Double Q-learning is more effective, it does not entirely eliminate the overestimation. We further show this reduction is not sufficient experimentally in Section \ref{sec:results}. 

As $\pi_{\phi_1}$ optimizes with respect to $Q_{\theta_1}$, using an independent estimate in the target update of $Q_{\theta_1}$ would avoid the bias introduced by the policy update. 
However the critics are not entirely independent, due to the use of the opposite critic in the learning targets, as well as the same replay buffer. As a result, for some states $s$ we will have $Q_{\theta_2}(s, \pi_{\phi_1}(s)) > Q_{\theta_1}(s, \pi_{\phi_1}(s))$. 
% While $Q_{\theta_2}(s, \pi_{\phi_1}(s))$ is a less biased estimate of $Q^\pi(s, \pi_{\phi_1}(s))$, for some states $s$ we will occasionally have $Q_{\theta_2}(s, \pi_{\phi_1}(s)) > Q_{\theta_1}(s, \pi_{\phi_1}(s))$. %%%%%%%%%%%%%%%%%% HERE? 
This is problematic because $Q_{\theta_1}(s, \pi_{\phi_1}(s))$ will generally overestimate the true value, and in certain areas of the state space the overestimation will be further exaggerated.  
%$\E \lb Q_{\theta_1}(s, \pi_{\phi_1}(s)) \rb > \E \lb Q^\pi(s, \pi_{\phi_1}(s)) \rb$, meaning that
To address this problem, we propose to simply upper-bound the less biased value estimate $Q_{\theta_2}$ by the biased estimate $Q_{\theta_1}$. This results in taking the minimum between the two estimates, to give the target update of our Clipped Double Q-learning algorithm: 
\begin{equation}
y_1 = r + \y \min_{i=1,2} Q_{\theta'_i}(s', \pi_{\phi_1}(s')).
\label{eq:clipped}
\end{equation}
% Intuitively, $Q_{\theta_2}$ acts as an unbiased estimate, while $Q_{\theta_1}$ acts as an approximate upper-bound to the value estimate. 
% With Clipped Double Q-learning, our worst case becomes the standard Q-learning target and we restrict any possible overestimation introduced into the value estimate. 
With Clipped Double Q-learning, the value target cannot introduce any additional overestimation over using the standard Q-learning target. 
While this update rule may induce an underestimation bias, this is far preferable to overestimation bias, as unlike overestimated actions, the value of underestimated actions will not be explicitly propagated through the policy update. 

%Note that overestimation due to function approximation is not a good measure of uncertainty, thus, using an over-estimating critic is not an effective way to implement optimism in the face of uncertainty \cite{kaelbling1996reinforcement}, and can lead to suboptimal policies \cite{thrun1993bias}. 

In implementation, computational costs can be reduced by using a single actor optimized with respect to $Q_{\theta_1}$. We then use the same target $y_2=y_1$ for $Q_{\theta_2}$. If $Q_{\theta_2} > Q_{\theta_1}$ then the update is identical to the standard update and induces no additional bias. If $Q_{\theta_2} < Q_{\theta_1}$, this suggests overestimation has occurred and the value is reduced similar to Double Q-learning. A proof of convergence in the finite MDP setting follows from this intuition. We provide formal details and justification in the supplementary material.

A secondary benefit is that by treating the function approximation error as a random variable we can see that the minimum operator should provide higher value to states with lower variance estimation error, as the expected minimum of a set of random variables decreases as the variance of the random variables increases. This effect means that the minimization in Equation~(\ref{eq:clipped}) will lead to a preference for states with low-variance value estimates, leading to safer policy updates with stable learning targets.

%%%%%%%%%%%%%%%%%%%%%%%%%%%%%%%%%%%%%%%%%
\section{Addressing Variance} \label{sec:var}
%%%%%%%%%%%%%%%%%%%%%%%%%%%%%%%%%%%%%%%%%

While Section \ref{sec:over} deals with the contribution of variance to overestimation bias, we also argue that variance itself should be directly addressed. Besides the impact on overestimation bias, high variance estimates provide a noisy gradient for the policy update. This is known to reduce learning speed \citep{sutton1998reinforcement} as well as hurt performance in practice. In this section we emphasize the importance of minimizing error at each update, build the connection between target networks and estimation error and propose modifications to the learning procedure of actor-critic for variance reduction. 

%%%%%%%%%%%%%%%%%%%%%%%%%%%%%%%%%%%%%%%%%
\subsection{Accumulating Error} \label{sec:acerror}
%%%%%%%%%%%%%%%%%%%%%%%%%%%%%%%%%%%%%%%%%

Due to the temporal difference update, where an estimate of the value function is built from an estimate of a subsequent state, there is a build up of error. While it is reasonable to expect small error for an individual update, these estimation errors can accumulate, resulting in the potential for large overestimation bias and suboptimal policy updates. This is exacerbated in a function approximation setting where the Bellman equation is never exactly satisfied, and each update leaves some amount of residual TD-error $\delta(s,a)$:
\begin{equation} \label{eqn:residual}
Q_\theta(s, a) = r + \y \E[Q_\theta(s', a')] - \delta(s,a).
\end{equation} 
It can then be shown that rather than learning an estimate of the expected return, the value estimate approximates the expected return minus the expected discounted sum of future TD-errors:
\begin{align}
&Q_\theta(s_t, a_t) = r_t + \y  \E [ Q_\theta(s_{t+1}, a_{t+1}) ] - \delta_t \nonumber \\
&= r_t + \y \E \lb r_{t+1} + \y \E \lb Q_\theta(s_{t+2},a_{t+2}) - \delta_{t+1} \rb \rb  - \delta_t \nonumber \\
%&...\\
% &= \E_{s \sim p_\pi} \lb \sum_{i=t}^T \y^{i-t} r_i 
% %\rb - \E_{s \sim p_\pi} \lb 
% - \sum_{i=t}^T \y^{i-t} \delta_i \rb.
&= \E_{s_i \sim p_\pi, a_i \sim \pi} \lb \sum_{i=t}^T \y^{i-t} (r_i - \delta_i) \rb.
\end{align}
If the value estimate is a function of future reward and estimation error, it follows that the variance of the estimate will be proportional to the variance of future reward and estimation error. Given a large discount factor $\y$, the variance can grow rapidly with each update if the error from each update is not tamed. Furthermore each gradient update only reduces error with respect to a small mini-batch which gives no guarantees about the size of errors in value estimates outside the mini-batch. 

%%%%%%%%%%%%%%%%%%%%%%%%%%%%%%%%%%%%%%%%%
\subsection{Target Networks and Delayed Policy Updates} \label{sec:dp}
%%%%%%%%%%%%%%%%%%%%%%%%%%%%%%%%%%%%%%%%%

In this section we examine the relationship between target networks and function approximation error, and show the use of a stable target reduces the growth of error. This insight allows us to consider the interplay between high variance estimates and policy performance, when designing reinforcement learning algorithms. 
 
Target networks are a well-known tool to achieve stability in deep reinforcement learning. As deep function approximators require multiple gradient updates to converge, target networks provide a stable objective in the learning procedure, and allow a greater coverage of the training data. Without a fixed target, each update may leave residual error which will begin to accumulate. While the accumulation of error can be detrimental in itself, when paired with a policy maximizing over the value estimate, it can result in wildly divergent values. 

\begin{figure} 
\centering
\captionsetup[subfloat]{captionskip=-8pt}
\includegraphics[width=\linewidth]{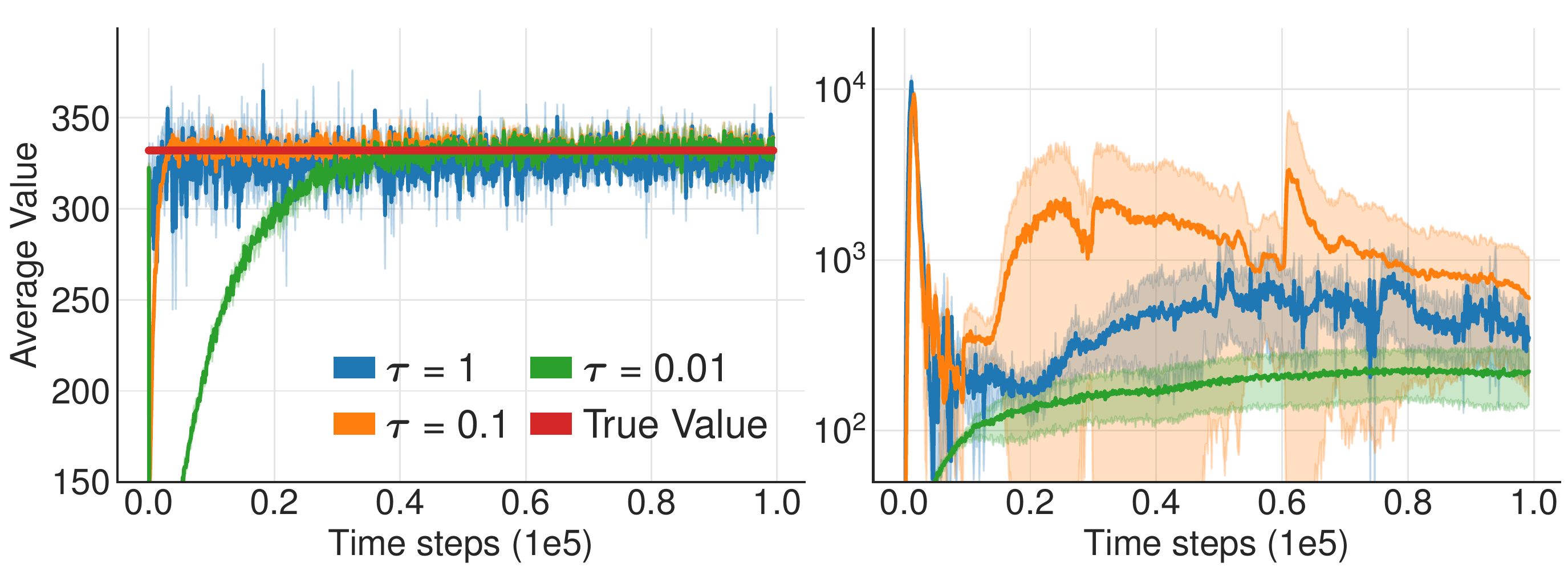}
\subfloat[Fixed Policy]{\hspace{0.56\linewidth}}
\subfloat[Learned Policy]{\hspace{0.44\linewidth}}
\caption{Average estimated value of a randomly selected state on Hopper-v1 without target networks, ($\tau=1$), and with slow-updating target networks, ($\tau=0.1, 0.01$), with a fixed and a learned policy.}
\label{fig:target}
\end{figure}

To provide some intuition, we examine the learning behavior with and without target networks on both the critic and actor in Figure \ref{fig:target}, where we graph the value, in a similar manner to Figure \ref{fig:overestimation}, in the Hopper-v1 environment. In (a) we compare the behavior with a fixed policy and in (b) we examine the value estimates with a policy that continues to learn, trained with the current value estimate. The target networks use a slow-moving update rate, parametrized by $\tau$. 

While updating the value estimate without target networks ($\tau=1$) increases the volatility, all update rates result in similar convergent behaviors when considering a fixed policy. However, when the policy is trained with the current value estimate, the use of fast-updating target networks results in highly divergent behavior. 

\textbf{When do actor-critic methods fail to learn?}
These results suggest that the divergence that occurs without target networks is the result of policy updates with a high variance value estimate. Figure \ref{fig:target}, as well as Section \ref{sec:over}, suggest failure can occur due to the interplay between the actor and critic updates. Value estimates diverge through overestimation when the policy is poor, and the policy will become poor if the value estimate itself is inaccurate. 

If target networks can be used to reduce the error over multiple updates, and policy updates on high-error states cause divergent behavior, then the policy network should be updated at a lower frequency than the value network, to first minimize error before introducing a policy update.
We propose delaying policy updates until the value error is as small as possible. 
% Using a slow-moving target network helps avoid jumps in error, but we propose to delay the policy updates until the value error is as small as possible. 
The modification is to only update the policy and target networks after a fixed number of updates $d$ to the critic. To ensure the TD-error remains small, we update the target networks slowly $\theta' \leftarrow \tau \theta + (1 - \tau) \theta'$. 

By sufficiently delaying the policy updates we limit the likelihood of repeating updates with respect to an unchanged critic. The less frequent policy updates that do occur will use a value estimate with lower variance, and in principle, should result in higher quality policy updates. This creates a two-timescale algorithm, as often required for convergence in the linear setting \cite{konda2003onactor}. The effectiveness of this strategy is captured by our empirical results presented in Section \ref{sec:results}, which show an improvement in performance while using fewer policy updates. 

%%%%%%%%%%%%%%%%%%%%%%%%%%%%%%%%%%%%%%%%%
\subsection{Target Policy Smoothing Regularization} \label{sec:TPN}
%%%%%%%%%%%%%%%%%%%%%%%%%%%%%%%%%%%%%%%%%

A concern with deterministic policies is they can overfit to narrow peaks in the value estimate. When updating the critic, a learning target using a deterministic policy is highly susceptible to inaccuracies induced by function approximation error, increasing the variance of the target. This induced variance can be reduced through regularization. We introduce a regularization strategy for deep value learning, target policy smoothing, which mimics the learning update from SARSA \cite{sutton1998reinforcement}. Our approach enforces the notion that similar actions should have similar value. While the function approximation does this implicitly, the relationship between similar actions can be forced explicitly by modifying the training procedure. We propose that fitting the value of a small area around the target action
\begin{equation}
y = r + \E_{\e} \lb Q_{\theta'}(s',  \pi_{\phi'}(s') + \e) \rb,
\end{equation}
would have the benefit of smoothing the value estimate by bootstrapping off of similar state-action value estimates. In practice, we can approximate this expectation over actions by adding a small amount of random noise to the target policy and averaging over mini-batches. This makes our modified target update: 
\begin{equation}
\begin{aligned}
y &= r + \y Q_{\theta'}(s', \pi_{\phi'}(s') + \e), \\
& \e \sim \clip(\N(0, \s), -c, c),
\end{aligned}
\end{equation}
where the added noise is clipped to keep the target close to the original action. 
The outcome is an algorithm reminiscent of Expected SARSA \cite{van2009theoretical}, where the value estimate is instead learned off-policy and the noise added to the target policy is chosen independently of the exploration policy.
The value estimate learned is with respect to a noisy policy defined by the parameter $\s$. 

Intuitively, it is known that policies derived from SARSA value estimates tend to be safer, as they provide higher value to actions resistant to perturbations. Thus, this style of update can additionally lead to improvement in stochastic domains with failure cases. A similar idea was introduced concurrently by \citet{Smoothie}, smoothing over $Q_\theta$, rather than $Q_{\theta'}$. 

%%%%%%%%%%%%%%%%%%%%%%%%%%%%%%%%%%%%%%%%%
\section{Experiments}
%%%%%%%%%%%%%%%%%%%%%%%%%%%%%%%%%%%%%%%%%

We present the Twin Delayed Deep Deterministic policy gradient algorithm (TD3), which builds on the Deep Deterministic Policy Gradient algorithm (DDPG) \cite{DDPG} by applying the modifications described in Sections \ref{sec:cdq}, \ref{sec:dp} and \ref{sec:TPN} to increase the stability and performance with consideration of function approximation error. 
TD3 maintains a pair of critics along with a single actor. For each time step, we update the pair of critics towards the minimum target value of actions selected by the target policy:
\begin{equation}
\begin{aligned}
y &= r + \y \min_{i=1,2} Q_{\theta'_i}(s', \pi_{\phi'}(s') + \e), \\
& \e \sim \clip(\N(0, \s), -c, c). \\
\end{aligned}
\end{equation}
Every $d$ iterations, the policy is updated with respect to $ Q_{\theta_1}$ following the deterministic policy gradient algorithm \cite{DPG}. TD3 is summarized in Algorithm \ref{alg:TD3}.

\begin{algorithm}[tb]
   \caption{TD3}
   \label{alg:TD3}
\begin{algorithmic}
%    \INPUT Horizon $T$, batch size $N$, exploration noise $\s_\textnormal{exploration}$, target network update rate $\tau$, target policy noise $\s_\textnormal{target}$, policy delay $d$. 
%    \STATE 
   \STATE Initialize critic networks $Q_{\theta_1}$, $Q_{\theta_2}$, and actor network $\pi_\phi$ with random parameters $\theta_1$, $\theta_2$, $\phi$
   \STATE Initialize target networks $\theta'_1 \leftarrow \theta_1$, $\theta'_2 \leftarrow \theta_2$, $\phi' \leftarrow \phi$
   \STATE Initialize replay buffer $\B$
   \FOR{$t=1$ {\bfseries to} $T$}
   \STATE Select action with exploration noise $a \sim \pi_\phi(s) + \e$, 
   \STATE $\e \sim \N(0, \sigma)$ and observe reward $r$ and new state $s'$
   \STATE Store transition tuple $(s, a, r, s')$ in $\B$ 
   \STATE 
   \STATE Sample mini-batch of $N$ transitions $(s, a, r, s')$ from $\B$
   %\STATE Select action with target policy noise: 
   \STATE $\tilde a \leftarrow \pi_{\phi'}(s') + \e, \quad \e \sim \clip(\N(0, \tilde \s), -c, c)$
   %\STATE Use Clipped Double Q-learning target:
   \STATE $y \leftarrow r + \y \min_{i=1,2} Q_{\theta'_i}(s', \tilde a)$
   %\STATE Update $\theta_i$ to minimize $N^{-1} \sum (y - Q_{\theta_i}(s,a))^2$
   %$\{Q_{\theta_i}\}_{i=1}^2$
   \STATE Update critics $\theta_i \leftarrow \argmin_{\theta_i} N^{-1} \sum (y - Q_{\theta_i}(s,a))^2$
   \IF{$t$ mod $d$}
   \STATE Update $\phi$ by the deterministic policy gradient:
   \STATE $\nabla_{\phi} J(\phi) = N^{-1} \sum \nabla_{a} Q_{\theta_1}(s, a) |_{a=\pi_{\phi}(s)} \nabla_{\phi} \pi_\phi(s)$
   \STATE Update target networks:
   \STATE $\theta'_i \leftarrow \tau \theta_i + (1 - \tau) \theta'_i$
   \STATE $\phi' \leftarrow \tau \phi + (1 - \tau) \phi'$
   \ENDIF
   \ENDFOR
\end{algorithmic}
\end{algorithm}

\begin{figure}[t] 
\centering
\subfloat[]{\includegraphics[width=0.25\linewidth]{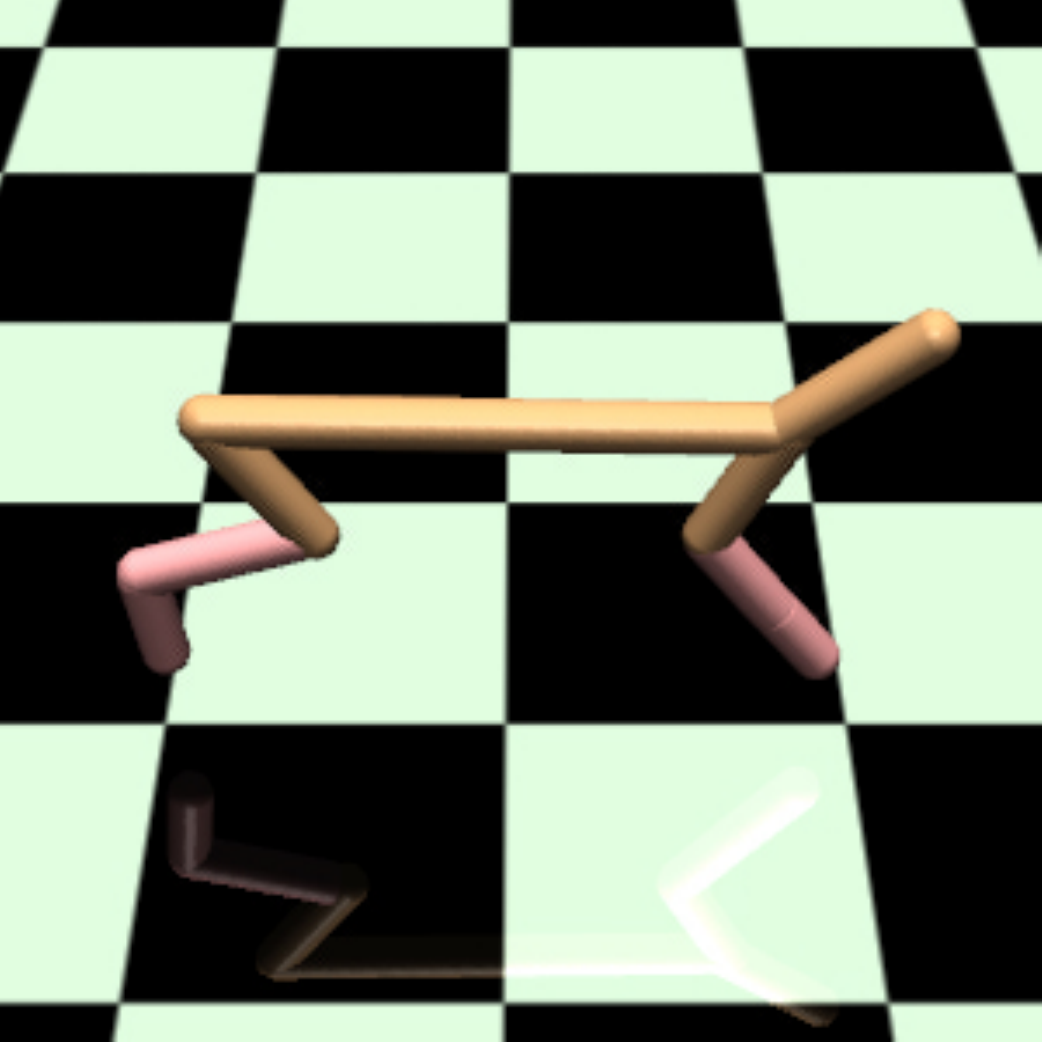}}
\subfloat[]{\includegraphics[width=0.25\linewidth]{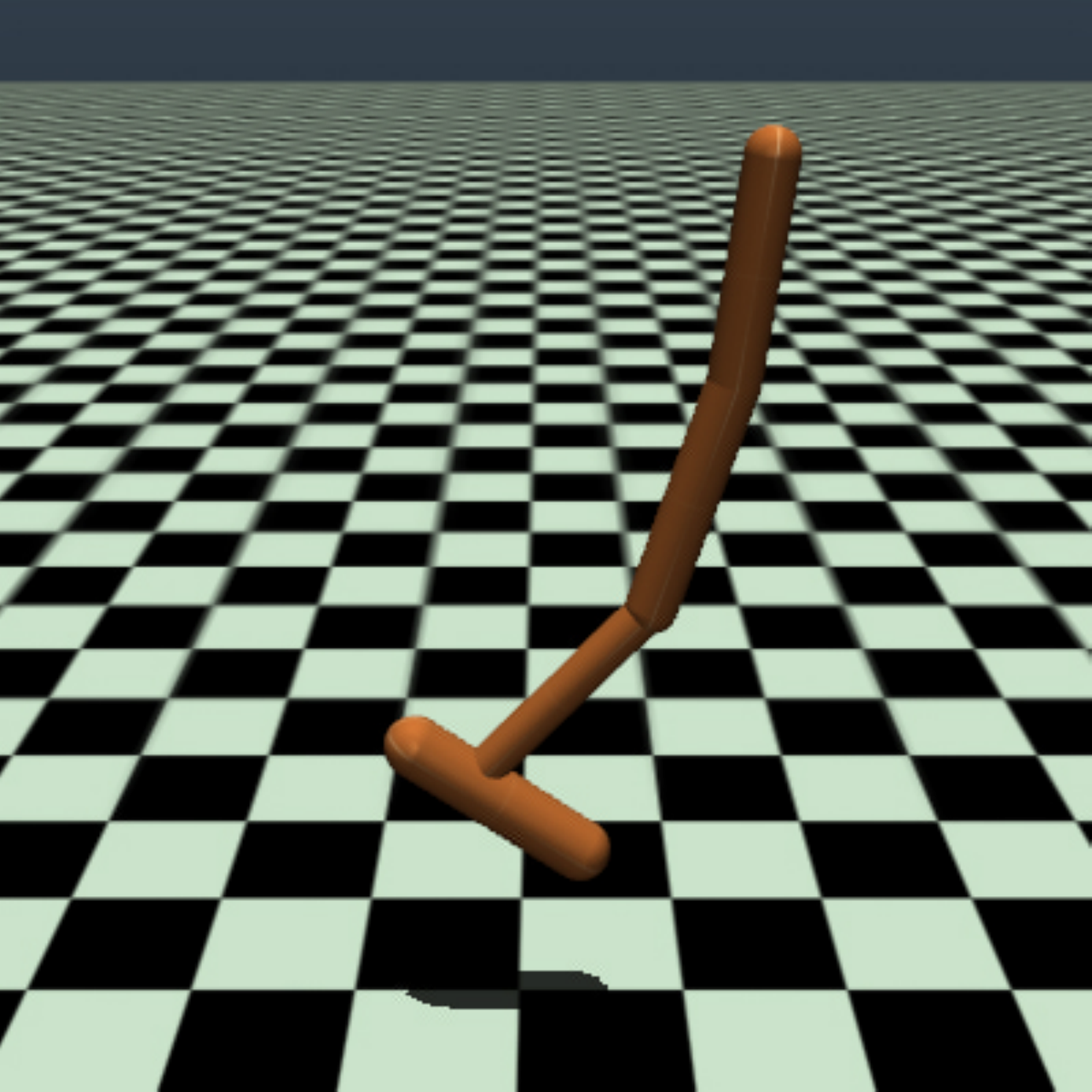}}
\subfloat[]{\includegraphics[width=0.25\linewidth]{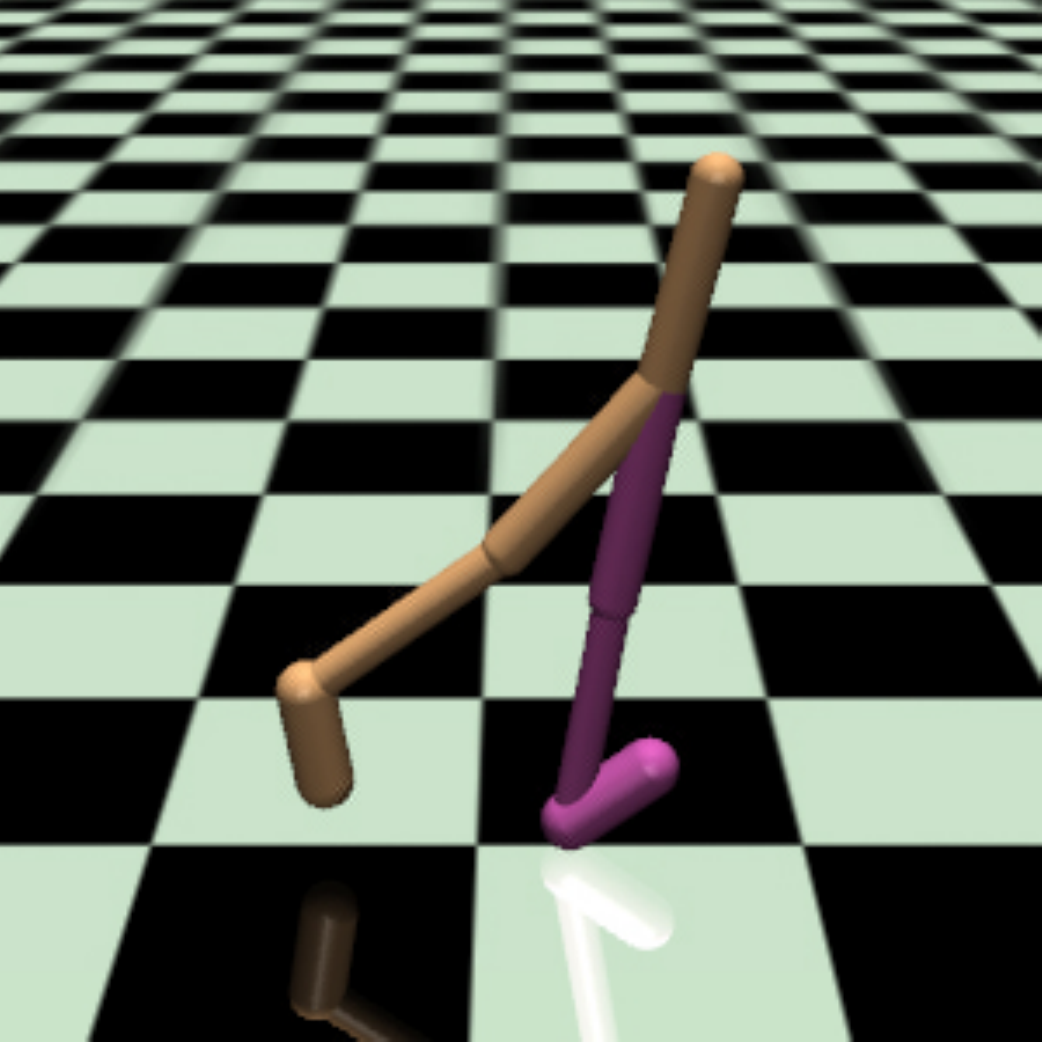}}
\subfloat[]{\includegraphics[width=0.25\linewidth]{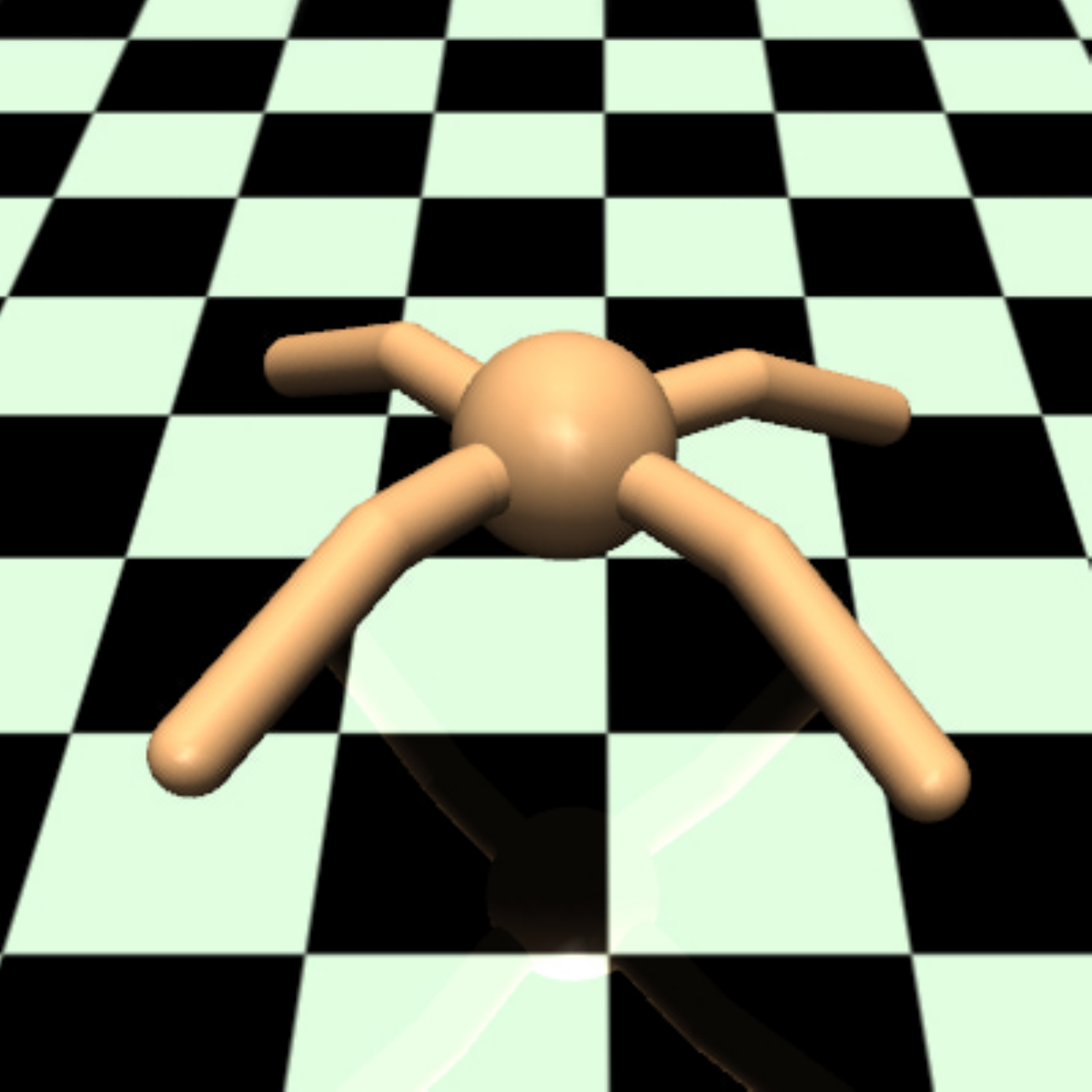}}
\caption{Example MuJoCo environments (a) HalfCheetah-v1, (b) Hopper-v1, (c) Walker2d-v1, (d) Ant-v1.}
\label{fig:env}
\end{figure}

%%%%%%%%%%%%%%%%%%%%%%%%%%%%%%%%%%%%%%%%%
\subsection{Evaluation} \label{sec:results}
%%%%%%%%%%%%%%%%%%%%%%%%%%%%%%%%%%%%%%%%%
 
\begin{figure*}[t]
\centering
\captionsetup[subfloat]{captionskip=-8pt}
\includegraphics[width=\linewidth]{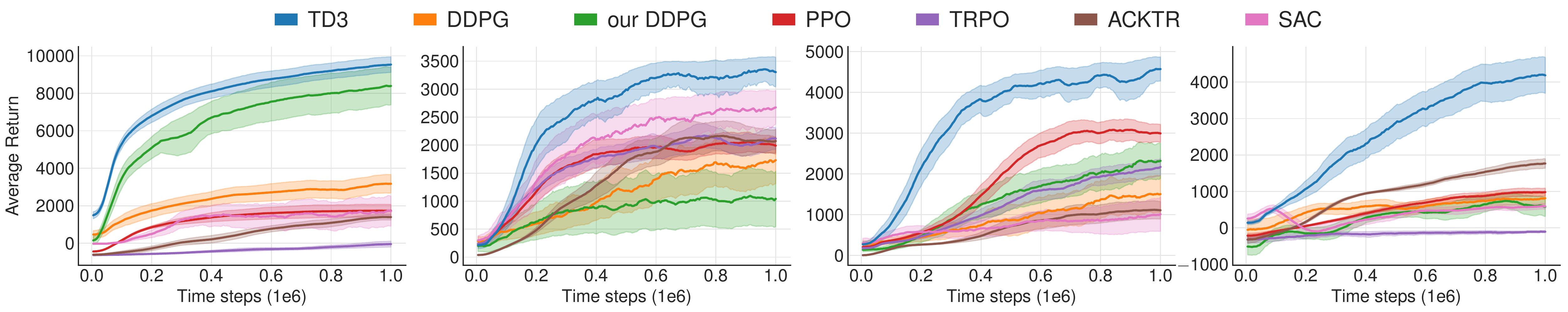}\\
\subfloat[HalfCheetah-v1]{\hspace{0.28\linewidth}}
\subfloat[Hopper-v1]{\hspace{0.24\linewidth}}
\subfloat[Walker2d-v1]{\hspace{0.24\linewidth}}
\subfloat[Ant-v1]{\hspace{0.24\linewidth}}\\
\includegraphics[trim={-6cm 0 6cm 1cm},clip, width=\linewidth]{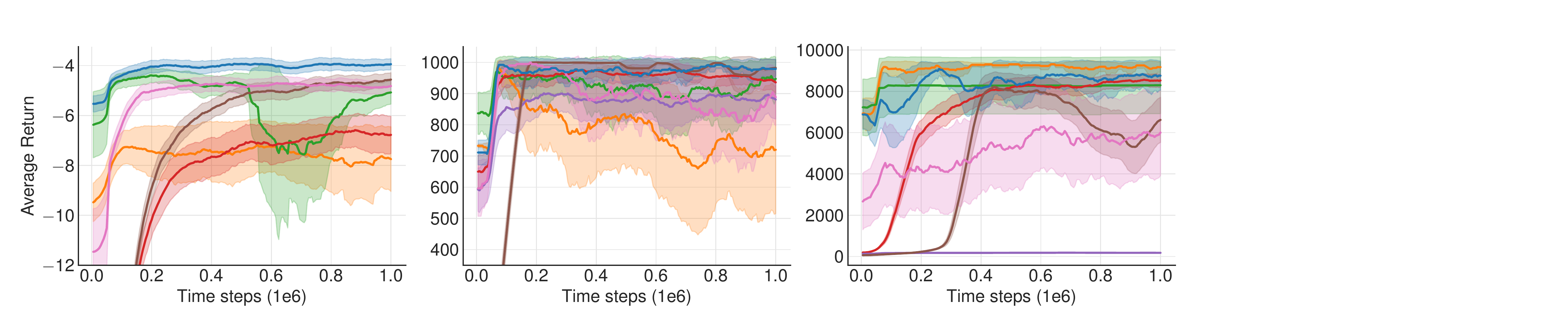}\\
\subfloat[Reacher-v1]{\hspace{0.28\linewidth}}
\subfloat[InvertedPendulum-v1]{\hspace{0.24\linewidth}}
\subfloat[InvertedDoublePendulum-v1]{\hspace{0.24\linewidth}}
\caption{Learning curves for the OpenAI gym continuous control tasks. The shaded region represents half a standard deviation of the average evaluation over 10 trials. Curves are smoothed uniformly for visual clarity.} %Some graphs are cropped to display the interesting regions.
\label{OpenAIcurves}
\end{figure*}

\begin{table*}[t]
\centering
\caption{Max Average Return over 10 trials of 1 million time steps. Maximum value for each task is bolded. $\pm$ corresponds to a single standard deviation over trials. }
\label{results}
%\vskip 0.15in
\begin{center}
\begin{small}
\begin{tabular}{lccccccc}
\toprule
\bf{Environment} & \bf{TD3} & \bf{DDPG} & \bf{Our DDPG} & \bf{PPO} & \bf{TRPO} & \bf{ACKTR} & \bf{SAC} \\
\midrule
HalfCheetah 	& \bf{9636.95 $\pm$ 859.065}& 3305.60 		& 8577.29 	& 1795.43 		& -15.57  & 1450.46 & 2347.19 \\  
Hopper 			& \bf{3564.07 $\pm$ 114.74} & 2020.46 		& 1860.02		& 2164.70		& 2471.30 & 2428.39 & 2996.66 \\
Walker2d 		& \bf{4682.82 $\pm$ 539.64} & 1843.85		& 3098.11 		& 3317.69		& 2321.47 & 1216.70 & 1283.67 \\
Ant 			& \bf{4372.44 $\pm$ 1000.33} & 1005.30 		& 888.77 		& 1083.20		& -75.85  & 1821.94 & 655.35 \\
Reacher 		& \bf{-3.60 $\pm$ 0.56}		& -6.51 	   	& \bf{-4.01} 	& -6.18			& -111.43 & -4.26 & -4.44\\
InvPendulum 	& \bf{1000.00 $\pm$ 0.00} 		& \bf{1000.00} 	& \bf{1000.00}  & \bf{1000.00}	& 985.40  & \bf{1000.00} & \bf{1000.00} \\
InvDoublePendulum 	& \bf{9337.47 $\pm$ 14.96} 	& \bf{9355.52} 	& 8369.95 		& 8977.94		& 205.85  & 9081.92 & 8487.15 \\
\bottomrule
\end{tabular}
\end{small}
\end{center}
\vskip -0.1in
\end{table*}

To evaluate our algorithm, we measure its performance on the suite of MuJoCo continuous control tasks \cite{mujoco}, interfaced through OpenAI Gym \cite{OpenAIGym} (Figure \ref{fig:env}). To allow for reproducible comparison, we use the original set of tasks from \citet{OpenAIGym} with no modifications to the environment or reward. 

For our implementation of DDPG \cite{DDPG}, we use a two layer feedforward neural network of 400 and 300 hidden nodes respectively, with rectified linear units (ReLU) between each layer for both the actor and critic, and a final tanh unit following the output of the actor. Unlike the original DDPG, the critic receives both the state and action as input to the first layer. Both network parameters are updated using Adam \cite{adam} with a learning rate of $10^{-3}$. After each time step, the networks are trained with a mini-batch of a 100 transitions, sampled uniformly from a replay buffer containing the entire history of the~agent. %without L2 regularization

The target policy smoothing is implemented by adding $\e \sim \N(0, 0.2)$ to the actions chosen by the target actor network, clipped to $(-0.5, 0.5)$, delayed policy updates consists of only updating the actor and target critic network every $d$ iterations, with $d=2$. While a larger $d$ would result in a larger benefit with respect to accumulating errors, for fair comparison, the critics are only trained once per time step, and training the actor for too few iterations would cripple learning. Both target networks are updated with $\tau=0.005$. 

To remove the dependency on the initial parameters of the policy we use a purely exploratory policy for the first 10000 time steps of stable length environments (HalfCheetah-v1 and Ant-v1) and the first 1000 time steps for the remaining environments. Afterwards, we use an off-policy exploration strategy, adding Gaussian noise $\mathcal{N}(0, 0.1)$ to each action. Unlike the original implementation of DDPG, we used uncorrelated noise for exploration as we found noise drawn from the Ornstein-Uhlenbeck \cite{uhlenbeck1930theory} process offered no performance benefits.

Each task is run for 1 million time steps with evaluations every 5000 time steps, where each evaluation reports the average reward over 10 episodes with no exploration noise. Our results are reported over 10 random seeds of the Gym simulator and the network initialization. 

We compare our algorithm against DDPG \cite{DDPG} as well as the state of art policy gradient algorithms: PPO \cite{ppo}, ACKTR \cite{acktr} and TRPO \cite{trpo}, as implemented by OpenAI's baselines repository \cite{baselines}, and SAC \cite{haarnoja2018soft}, as implemented by the author's GitHub\footnote{See the supplementary material for hyper-parameters and a discussion on the discrepancy in the reported results of SAC.}. 
%For fair comparison with our method, we train SAC once per time step, rather than 4 times, which explains the discrepancy with their reported results\footnote{In a recent version, Soft Actor-Critic authors updated their algorithm to include Clipped Double Q-learning, which was not included in this implementation.}. 
Additionally, we compare our method with our re-tuned version of DDPG, which includes all architecture and hyper-parameter modifications to DDPG without any of our proposed adjustments. 
A full comparison between our re-tuned version and the baselines DDPG is provided in the supplementary material.

Our results are presented in Table \ref{results} and learning curves in Figure \ref{OpenAIcurves}. TD3 matches or outperforms all other algorithms in both final performance and learning speed across all tasks. 

%%%%%%%%%%%%%%%%%%%%%%%%%%%%%%%%%%%%%%%%%
\subsection{Ablation Studies}
%%%%%%%%%%%%%%%%%%%%%%%%%%%%%%%%%%%%%%%%%

We perform ablation studies to understand the contribution of each individual component: Clipped Double Q-learning (Section \ref{sec:cdq}), delayed policy updates (Section \ref{sec:dp}) and target policy smoothing (Section \ref{sec:TPN}). We present our results in Table \ref{table:ablation} in which we compare the performance of removing each component from TD3 along with our modifications to the architecture and hyper-parameters. Additional learning curves can be found in the supplementary material. 

The significance of each component varies task to task. While the addition of only a single component causes insignificant improvement in most cases, the addition of combinations performs at a much higher level.  The full algorithm outperforms every other combination in most tasks. Although the actor is trained for only half the number of iterations, the inclusion of delayed policy update generally improves performance, while reducing training time. 

\begin{table}[ht]
\centering
\caption{Average return over the last 10 evaluations over 10 trials of 1 million time steps, comparing ablation over delayed policy updates (DP), target policy smoothing (TPS), Clipped Double Q-learning (CDQ) and our architecture, hyper-parameters and exploration (AHE). Maximum value for each task is bolded.}
\label{table:ablation}
%\vskip 0.15in
\begin{center}
\begin{small}
\begin{tabular}{lccccr}
\toprule
\bf{Method} & \bf{HCheetah}&\bf{Hopper}	& \bf{Walker2d} & \bf{Ant} \\
\midrule
TD3 		& 9532.99 		& \bf{3304.75} 	& \bf{4565.24} 	& \bf{4185.06} \\
DDPG		& 3162.50 		& 1731.94 		& 1520.90 		& 816.35 \\
AHE			& 8401.02 		& 1061.77 		& 2362.13 		& 564.07 \\
\midrule
AHE + DP 	& 7588.64 		& 1465.11 		& 2459.53 		& 896.13 \\
AHE + TPS 	& 9023.40 		& 907.56 		& 2961.36 		& 872.17 \\
AHE + CDQ 	& 6470.20 		& 1134.14 		& 3979.21 		& 3818.71 \\
\midrule
TD3 - DP 	& 9590.65		& 2407.42 		& \bf{4695.50} 	& 3754.26 \\
TD3 - TPS 	& 8987.69 		& 2392.59 		& 4033.67 		& \bf{4155.24} \\
TD3 - CDQ 	& 9792.80	  	& 1837.32 		& 2579.39 		& 849.75 \\
\midrule
DQ-AC 		& 9433.87		& 1773.71		& 3100.45 		& 2445.97 \\
DDQN-AC 	& \bf{10306.90}	& 2155.75 		& 3116.81 		& 1092.18 \\
\bottomrule
\end{tabular}
\end{small}
\end{center}
\vskip -0.1in
\end{table}

We additionally compare the effectiveness of the actor-critic variants of Double Q-learning \cite{hasselt2010double} and Double DQN \cite{DoubleDQN}, denoted DQ-AC and DDQN-AC respectively, in Table \ref{table:ablation}. For fairness in comparison, these methods also benefited from delayed policy updates, target policy smoothing and use our architecture and hyper-parameters. Both methods were shown to reduce overestimation bias less than Clipped Double Q-learning in Section \ref{sec:over}. This is reflected empirically, as both methods result in insignificant improvements over TD3 - CDQ, with an exception in the Ant-v1 environment, which appears to benefit greatly from any overestimation reduction. As the inclusion of Clipped Double Q-learning into our full method outperforms both prior methods, this suggests that subduing the overestimations from the unbiased estimator is an effective measure to improve performance. 

%%%%%%%%%%%%%%%%%%%%%%%%%%%%%%%%%%%%%%%%%
\section{Conclusion}
%%%%%%%%%%%%%%%%%%%%%%%%%%%%%%%%%%%%%%%%%

Overestimation has been identified as a key problem in value-based methods. In this paper, we establish overestimation bias is also problematic in actor-critic methods. We find the common solutions for reducing overestimation bias in deep Q-learning with discrete actions are ineffective in an actor-critic setting, and develop a novel variant of Double Q-learning which limits possible overestimation. Our results demonstrate that mitigating overestimation can greatly improve the performance of modern algorithms. 

Due to the connection between noise and overestimation, we examine the accumulation of errors from temporal difference learning. Our work investigates the importance of a standard technique in deep reinforcement learning, target networks, and examines their role in limiting errors from imprecise function approximation and stochastic optimization. Finally, we introduce a SARSA-style regularization technique which modifies the temporal difference target to bootstrap off similar state-action pairs. 

Taken together, these improvements define our proposed approach, the Twin Delayed Deep Deterministic policy gradient algorithm (TD3), which greatly improves both the learning speed and performance of DDPG in a number of challenging tasks in the continuous control setting. Our algorithm exceeds the performance of numerous state of the art algorithms. As our modifications are simple to implement, they can be easily added to any other actor-critic algorithm. 

\bibliography{example_paper}
\bibliographystyle{icml2018}

% % % %%%%%%%%%%%%%%%%%%%%%%%%%%%%%%%%%%%%%%%%%
\clearpage
\onecolumn

\icmltitle{Supplementary Material}
% \icmltitle{Addressing Function Approximation Error in Actor-Critic Methods: Supplementary Material}
% \icmltitlerunning{Addressing Function Approximation Error in Actor-Critic Methods: Supplementary Material}
%%%%%%%%%%%%%%%%%%%%%%%%%%%%%%%%%%%%%%%%%

%%%%%%%%%%%%%%%%%%%%%%%%%%%%%%%%%%%%%%%%%
\appendix
%%%%%%%%%%%%%%%%%%%%%%%%%%%%%%%%%%%%%%%%%

%%%%%%%%%%%%%%%%%%%%%%%%%%%%%%%%%%%%%%%%%
\section{Proof of Convergence of Clipped Double Q-Learning} \label{sec:sm_proof}
%%%%%%%%%%%%%%%%%%%%%%%%%%%%%%%%%%%%%%%%%

In a version of Clipped Double Q-learning for a finite MDP setting, we maintain two tabular value estimates $Q^A$, $Q^B$. At each time step we select actions $a^* = \argmax_a Q^A(s,a)$ and then perform an update by setting target $y$:
\begin{equation}
\begin{aligned}
a^* &= \argmax_a Q^A(s',a) \\
y &= r + \y \min(Q^A(s',a^*), Q^B(s',a^*)),
\end{aligned}
\end{equation}
and update the value estimates with respect to the target and learning rate $\alpha_t(s,a)$: 
\begin{equation}
\begin{aligned}
Q^A(s,a) &= Q^A(s,a) + \alpha_t(s,a)(y - Q^A(s,a)) \\
Q^B(s,a) &= Q^B(s,a) + \alpha_t(s,a)(y - Q^B(s,a)). 
\end{aligned}
\end{equation}

In a finite MDP setting, Double Q-learning is often used to deal with noise induced by random rewards or state transitions, and so either $Q^A$ or $Q^B$ is updated randomly. However, in a function approximation setting, the interest may be more towards the approximation error and thus we can update both $Q^A$ and $Q^B$ at each iteration. The proof extends naturally to updating either randomly. 

The proof borrows heavily from the proof of convergence of SARSA \cite{singh2000convergence} as well as Double Q-learning \cite{hasselt2010double}. The proof of lemma 1 can be found in \citet{singh2000convergence}, building on a proposition from \citet{bertsekas1995dynamic}. 

\textbf{Lemma 1.} Consider a stochastic process $(\zeta_t, \Delta_t, F_t), t \geq 0$ where $\zeta_t, \Delta_t, F_t: X \rightarrow \mathbb{R}$ satisfy the equation:
\begin{equation}
\Delta_{t+1}(x_t) = (1 - \zeta_t(x_t))\Delta_t(x_t) + \zeta_t(x_t)F_t(x_t),
\end{equation}
where $x_t \in X$ and $t=0,1,2,...$. Let $P_t$ be a sequence of increasing $\sigma$-fields such that $\zeta_0$ and $\Delta_0$ are $P_0$-measurable and $\zeta_t, \Delta_t$ and $F_{t-1}$ are $P_t$-measurable, $t=1,2,...$. Assume that the following hold:
\begin{enumerate}[topsep=0pt]
\item The set $X$ is finite.
\item $\zeta_t(x_t) \in [0,1], \sum_t \zeta_t(x_t) = \infty, \sum_t (\zeta_t(x_t))^2 < \infty$ with probability $1$ and $\forall x \neq x_t : \zeta(x) = 0$.
\item $|| \E \lb F_t | P_t \rb || \leq \kappa || \Delta_t || + c_t$ where $\kappa \in [0,1)$ and $c_t$ converges to $0$ with probability $1$.
\item Var$\lb F_t(x_t)|P_t \rb \leq K(1 + \kappa||\Delta_t||)^2$, where $K$ is some constant 
\end{enumerate}
Where $||\cdot||$ denotes the maximum norm. Then $\Delta_t$ converges to $0$ with probability $1$. 

\textbf{Theorem 1.} Given the following conditions:
\begin{enumerate}[topsep=0pt]
\item Each state action pair is sampled an infinite number of times. 
\item The MDP is finite. 
\item $\y \in [0,1)$.
\item Q values are stored in a lookup table.
\item Both $Q^A$ and $Q^B$ receive an infinite number of updates.
\item The learning rates satisfy $\alpha_t(s,a) \in [0,1], \sum_t \alpha_t(s,a) = \infty, \sum_t(\alpha_t(s,a))^2 < \infty$ with probability $1$ and $\alpha_t(s,a) = 0, \forall(s,a) \neq (s_t,a_t)$.
\item Var$\lb r(s,a) \rb < \infty, \forall s,a$. 
\end{enumerate}
Then Clipped Double Q-learning will converge to the optimal value function $Q^*$, as defined by the Bellman optimality equation, with probability $1$.

\textbf{Proof of Theorem 1.} We apply Lemma 1 with $P_t = \{Q^A_0, Q^B_0, s_0, a_0, \alpha_0, r_1, s_1, ..., s_t, a_t\}, X = S \times A, \Delta_t = Q^A_t - Q^*, \zeta_t = \alpha_t$.  

First note that condition 1 and 4 of the lemma holds by the conditions 2 and 7 of the theorem respectively. Lemma condition 2 holds by the theorem condition 6 along with our selection of $\zeta_t = \alpha_t$.  

Defining $a^* = \argmax_a Q^A(s_{t+1}, a)$ we have
\begin{equation}
\begin{aligned}
\Delta_{t+1}(s_t,a_t) &= (1 - \alpha_t(s_t,a_t))(Q^A_t(s_t,a_t) - Q^*(s_t,a_t)) \\
& \qquad + \alpha_t(s_t,a_t)(r_t + \y \min (Q^A_t(s_{t+1}, a^*), Q^B_t(s_{t+1}, a^*)) - Q^*(s_t,a_t)) \\ 
&= (1 - \alpha_t(s_t,a_t))\Delta_t(s_t,a_t) + \alpha_t(s_t,a_t)F_t(s_t,a_t)), 
\end{aligned}
\end{equation}
where we have defined $F_t(s_t,a_t)$ as:
\begin{equation}
\begin{aligned}
F_t(s_t, a_t) &= r_t + \y \min (Q^A_t(s_{t+1}, a^*), Q^B_t(s_{t+1}, a^*)) - Q^*_t(s_t,a_t) \\ 
&=r_t + \y \min (Q^A_t(s_{t+1}, a^*), Q^B_t(s_{t+1}, a^*)) - Q^*_t(s_t,a_t) + \y Q^A_t(s_{t+1}, a^*) - \y Q^A_t(s_{t+1}, a^*) \\
&= F^Q_t(s_t,a_t) + c_t,  
\end{aligned}
\end{equation}
where $F^Q_t = r_t + \y Q^A_t(s_{t+1}, a^*) - Q^*_t(s_t,a_t)$ denotes the value of $F_t$ under standard Q-learning and $c_t = \y \min (Q^A_t(s_{t+1}, a^*), Q^B_t(s_{t+1}, a^*)) - \y Q^A_t(s_{t+1}, a^*)$.
As $\E \lb F^Q_t | P_t \rb \leq \y ||\Delta_t||$ is a well-known result, then condition 3 of lemma 1 holds if it can be shown that $c_t$ converges to 0 with probability 1. 

Let $y = r_t + \y \min (Q^B_t(s_{t+1}, a^*), Q^A_t(s_{t+1}, a^*))$ and $\Delta^{BA}_t(s_t,a_t) = Q^B_t(s_t,a_t) - Q^A_t(s_t,a_t)$, where $c_t$ converges to 0 if $\Delta^{BA}$ converges to 0.
The update of $\Delta^{BA}_t$ at time $t$ is the sum of updates of $Q^A$ and $Q^B$:
\begin{equation}
\begin{aligned}
\Delta^{BA}_{t+1}(s_t,a_t) &= \Delta^{BA}_t(s_t,a_t) + \alpha_t(s_t,a_t) \lp y - Q^B_t(s_t,a_t) - (y - Q^A_t(s_t,a_t)) \rp \\ 
&= \Delta^{BA}_t(s_t,a_t) + \alpha_t(s_t,a_t) \lp Q^A_t(s_t,a_t) - Q^B_t(s_t,a_t) \rp \\ 
&= (1 - \alpha_t(s_t,a_t))\Delta^{BA}_t(s_t,a_t).
\end{aligned}
\end{equation}
Clearly $\Delta^{BA}_t$ will converge to $0$, which then shows we have satisfied condition 3 of lemma 1, implying that $Q^A(s_t,a_t)$ converges to $Q^*_t(s_t,a_t)$. Similarly, we get convergence of $Q^B(s_t,a_t)$ to the optimal vale function by choosing $\Delta_t = Q^B_t - Q^*$ and repeating the same arguments, thus proving theorem 1. 

%%%%%%%%%%%%%%%%%%%%%%%%%%%%%%%%%%%%%%%%%
\section{Overestimation Bias in Deterministic Policy Gradients} \label{sec:sm_over}
%%%%%%%%%%%%%%%%%%%%%%%%%%%%%%%%%%%%%%%%%

If the gradients from the deterministic policy gradient update are unnormalized, this overestimation is still guaranteed to occur under a slightly stronger condition on the expectation of the value estimate. Assume the approximate value function is equal to the true value function, in expectation over the steady-state distribution, with respect to policy parameters between the original policy and in the direction of the true policy update:
\begin{equation}
\begin{aligned}
\label{condition}
& \E_{s \sim \pi} \lb Q_\theta(s, \pi_\textnormal{new}(s))\rb = \E_{s \sim \pi} \lb Q^\pi(s, \pi_\textnormal{new}(s)) \rb \\
& \forall \phi_\textnormal{new} \in [\phi,\phi + \beta (\phi_{\textnormal{true}} - \phi)] \text{ such that } \beta > 0.
\end{aligned}
\end{equation}
Noting that $\phi_{\textnormal{true}}$ maximizes the rate of change of the true value $\Delta^\pi_{\textnormal{true}} = Q^\pi(s,\pi_\textnormal{true}(s)) - Q^\pi(s,\pi_\phi(s))$, $\Delta^\pi_{\textnormal{true}} \geq \Delta^\pi_{\textnormal{approx}}$. By the given condition \ref{condition} the maximal rate of change of the approximate value must be at least as great $\Delta^\theta_{\textnormal{approx}} \geq \Delta^\pi_{\textnormal{true}}$. Given $Q_\theta(s, \pi_\phi) = Q^\pi(s, \pi_\phi)$ this implies $Q_\theta(s,\pi_\textnormal{approx}(s)) \geq Q^\pi(s,\pi_\textnormal{true}(s)) \geq Q^\pi(s,\pi_\textnormal{approx}(s))$, showing an overestimation of the value function.
%\Rightarrow Q_\theta(s,\pi_\textnormal{approx}(s)) - Q_\theta(s,\pi_\phi(s)) \geq \Delta^\pi$. Summing this change with condition \ref{condition} then shows an overestimation of the value function. 

\clearpage
%%%%%%%%%%%%%%%%%%%%%%%%%%%%%%%%%%%%%%%%%
\section{DDPG Network and Hyper-parameter Comparison}
\label{sec:sm_hyp}
%%%%%%%%%%%%%%%%%%%%%%%%%%%%%%%%%%%%%%%%%

DDPG Critic Architecture
\begin{verbatim}
(state dim, 400)
ReLU
(action dim + 400, 300)
ReLU
(300, 1)
\end{verbatim}
DDPG Actor Architecture
\begin{verbatim}
(state dim, 400)
ReLU
(400, 300)
ReLU
(300, 1)
tanh
\end{verbatim}

Our Critic Architecture
\begin{verbatim}
(state dim + action dim, 400)
ReLU
(action dim + 400, 300)
RelU
(300, 1)
\end{verbatim}
Our Actor Architecture
\begin{verbatim}
(state dim, 400)
ReLU
(400, 300)
RelU
(300, 1)
tanh
\end{verbatim}

\begin{table}
\centering
\caption{A complete comparison of hyper-parameter choices between our DDPG and the OpenAI baselines implementation \cite{baselines}.}
\begin{center}
\begin{small}
\begin{tabular}{lcc}
\toprule
\bf{Hyper-parameter} & \bf{Ours} & \bf{DDPG} \\
\midrule
Critic Learning Rate & $10^{-3}$ & $10^{-3}$ \\
Critic Regularization & None & $10^{-2} \cdot ||\theta||^2$ \\
Actor Learning Rate & $10^{-3}$ & $10^{-4}$ \\
Actor Regularization & None & None \\
Optimizer & Adam & Adam \\
Target Update Rate ($\tau$) & $5 \cdot 10^{-3}$ & $10^{-3}$ \\ 
Batch Size & $100$ & $64$ \\ 
Iterations per time step & $1$ & $1$ \\
Discount Factor & $0.99$ & $0.99$ \\
Reward Scaling & $1.0$ & $1.0$ \\
Normalized Observations & False & True \\
Gradient Clipping & False & False \\
Exploration Policy & $\N(0, 0.1)$ & OU, $\theta=0.15, \mu=0, \sigma=0.2$ \\ 
\bottomrule
\end{tabular}
\end{small}
\end{center}
\end{table}

%%%%%%%%%%%%%%%%%%%%%%%%%%%%%%%%%%%%%%%%%
\section{Additional Implementation Details}
%%%%%%%%%%%%%%%%%%%%%%%%%%%%%%%%%%%%%%%%%

For clarity in presentation, certain implementation details were omitted, which we describe here. For the most complete possible description of the algorithm, code can be found on our GitHub (\url{https://github.com/sfujim/TD3}). 

Our implementation of both DDPG and TD3 follows a standard practice in deep Q-learning, in which the update differs for terminal transitions. For transitions where the episode terminates by reaching some failure state, and not due to the episode running until the max horizon, the value of $Q(s, \cdot)$ is set to $0$ in the target $y$: 
$$
y = 
  \begin{cases}
   r  & \text{if terminal } s' \text{ and } t < \text{max horizon} \\
   r + \y Q_{\theta'}(s',\pi_{\phi'}(s')) & \text{else} \\
  \end{cases}
$$

For target policy smoothing (Section \ref{sec:TPN}), the added noise is clipped to the range of possible actions, to avoid error introduced by using values of impossible actions: 
$$
\begin{aligned}
y &= r + \y Q_{\theta'}(s', \clip (\pi_{\phi'}(s') + \e, \text{min action}, \text{max action})), \\
& \e \sim \clip(\N(0, \s), -c, c).
\end{aligned}
$$

%
% done and terminal, and . . . episodic and . . . clipping 

%%%%%%%%%%%%%%%%%%%%%%%%%%%%%%%%%%%%%%%%%
\section{Soft Actor-Critic Implementation Details}
%%%%%%%%%%%%%%%%%%%%%%%%%%%%%%%%%%%%%%%%%

For our implementation of Soft Actor-Critic \cite{haarnoja2018soft} we use the code provided by the author (\url{https://github.com/haarnoja/sac}), using the hyper-parameters described by the paper. We use a Gaussian mixture policy with 4 Gaussian distributions, except for the Reacher-v1 task, where we use a single Gaussian distribution due to numerical instability issues in the provided implementation. We use the environment-dependent reward scaling as described by the authors, multiplying the rewards by 3 for Walker2d-v1 and Ant-v1, and 1 for all remaining environments. 

For fair comparison with our method, we train for only 1 iteration per time step, rather than the 4 iterations used by the results reported by the authors. This along with fewer total time steps should explain for the discrepancy in results on some of the environments. Additionally, we note this comparison is against a prior version of Soft Actor-Critic, while the most recent variant includes our Clipped Double Q-learning in the value update and produces competitive results to TD3 on most tasks. 

\clearpage

%%%%%%%%%%%%%%%%%%%%%%%%%%%%%%%%%%%%%%%%%
\section{Additional Learning Curves} \label{sec:sm_curves}
%%%%%%%%%%%%%%%%%%%%%%%%%%%%%%%%%%%%%%%%%

\begin{figure*}[h]
\centering
\captionsetup[subfloat]{captionskip=-10pt}
\includegraphics[width=\linewidth]{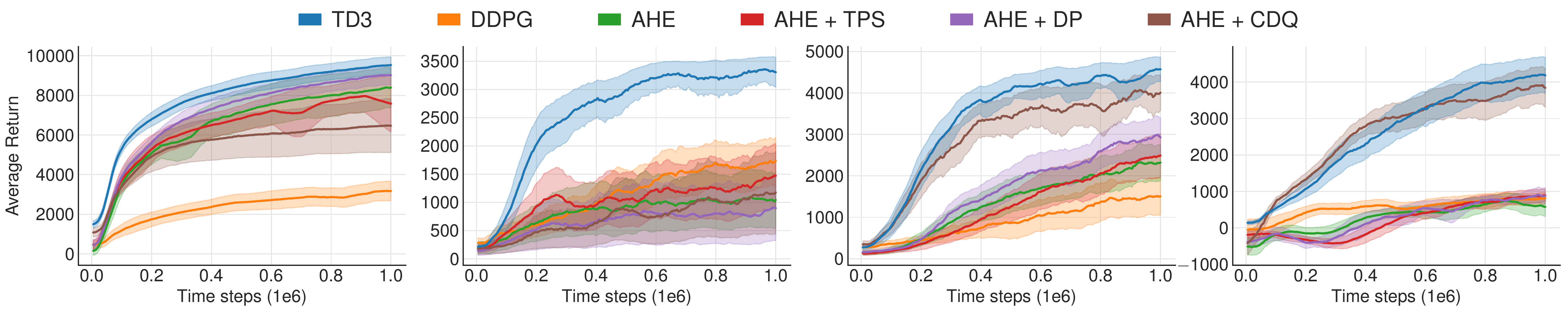}
\subfloat[HalfCheetah-v1]{\hspace{.28\linewidth}}
\subfloat[Hopper-v1]{\hspace{.24\linewidth}}
\subfloat[Walker2d-v1]{\hspace{.24\linewidth}}
\subfloat[Ant-v1]{\hspace{.24\linewidth}}
\caption{Ablation over the varying modifications to our DDPG (AHE), comparing the subtraction of delayed policy updates (TD3 - DP), target policy smoothing \mbox{(TD3 - TPS)} and Clipped Double Q-learning (TD3 - CDQ).}
\label{fig:ablation}
\end{figure*}

\begin{figure*}[h]
\centering
\captionsetup[subfloat]{captionskip=-10pt}
\includegraphics[width=\linewidth]{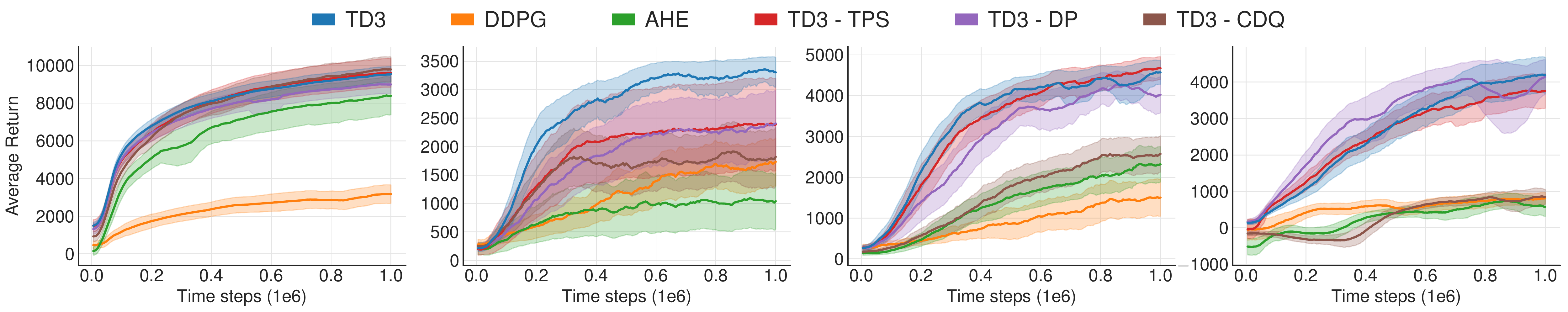}
\subfloat[HalfCheetah-v1]{\hspace{.28\linewidth}}
\subfloat[Hopper-v1]{\hspace{.24\linewidth}}
\subfloat[Walker2d-v1]{\hspace{.24\linewidth}}
\subfloat[Ant-v1]{\hspace{.24\linewidth}}
\caption{Ablation over the varying modifications to our DDPG (AHE), comparing the addition of delayed policy updates (AHE + DP), target policy smoothing (AHE + TPS) and Clipped Double Q-learning (AHE + CDQ).}
\label{fig:double_ablation}
\end{figure*}

\begin{figure*}[h]
\centering
\captionsetup[subfloat]{captionskip=-10pt}
\includegraphics[width=\linewidth]{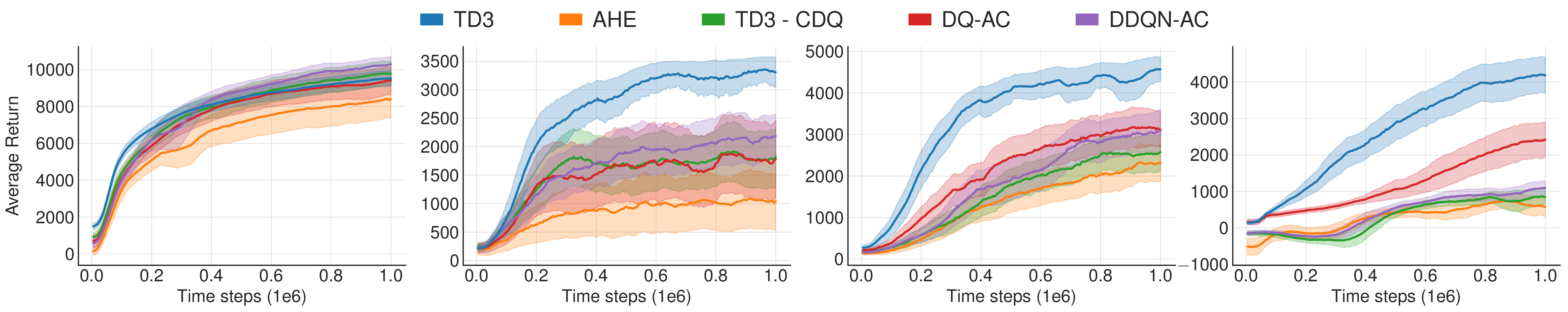}
\subfloat[HalfCheetah-v1]{\hspace{.28\linewidth}}
\subfloat[Hopper-v1]{\hspace{.24\linewidth}}
\subfloat[Walker2d-v1]{\hspace{.24\linewidth}}
\subfloat[Ant-v1]{\hspace{.24\linewidth}}
\caption{Comparison of TD3 and the Double Q-learning (DQ-AC) and Double DQN (DDQN-AC) actor-critic variants, which also leverage delayed policy updates and target policy smoothing.}
\label{fig:double}
\end{figure*}

% \clearpage

% \bibliography{example_paper}
% \bibliographystyle{icml2018}

\end{document}